\documentclass[10pt, a4paper]{article}
\usepackage{lrec2022} 
\usepackage{multibib}
\newcites{languageresource}{Language Resources}
\usepackage{graphicx}
\usepackage{tabularx}
\usepackage{soul}
\usepackage{url}
\usepackage{titlesec}
\titleformat{\section}{\normalfont\large\bf\center}{\thesection.}{1em}{}
\titleformat{\subsection}{\normalfont\SmallTitleFont\bf\raggedright}{\thesubsection.}{1em}{}
\titleformat{\subsubsection}{\normalfont\normalsize\bf\raggedright}{\thesubsubsection.}{1em}{}
\renewcommand\thesection{\arabic{section}}
\renewcommand\thesubsection{\thesection.\arabic{subsection}}
\renewcommand\thesubsubsection{\thesubsection.\arabic{subsubsection}}

\usepackage{epstopdf}
\usepackage[utf8]{inputenc}

\usepackage[colorlinks = true,
            linkcolor = blue,
            urlcolor  = blue,
            citecolor = black,
            anchorcolor = blue]{hyperref}
\usepackage{xstring}
\usepackage{caption}
\usepackage{subcaption}

\usepackage{color}
\usepackage[inline, shortlabels]{enumitem}

\title{MS-LaTTE: A Dataset of Where and When To-do Tasks are Completed}

\name{Sujay Kumar Jauhar, Nirupama Chandrasekaran, Michael Gamon, Ryen W. White}

\address{Microsoft Research \\
         Redmond, WA, USA \\
         \{sjauhar, niruc, mgamon, ryenw\}@microsoft.com\\}

\abstract{
Tasks are a fundamental unit of work in the daily lives of people, who are increasingly using digital means to keep track of, organize, triage and act on them. These digital tools -- such as task management applications -- provide a unique opportunity to study and understand tasks and their connection to the real world, and through intelligent assistance, help people be more productive. By logging signals such as text, timestamp information, and social connectivity graphs, an increasingly rich and detailed picture of how tasks are created and organized, what makes them important, and who acts on them, can be progressively developed. 
Yet the context around actual task completion remains fuzzy, due to the basic disconnect between actions taken in the real world and telemetry recorded in the digital world.
Thus, in this paper we compile and release a novel, real-life, large-scale dataset called MS-LaTTE that captures two core aspects of the context surrounding task completion: location and time. We describe our annotation framework and conduct a number of analyses on the data that were collected, demonstrating that it captures intuitive contextual properties for common tasks. Finally, we test the dataset on the two problems of predicting spatial and temporal task co-occurrence, concluding that predictors for co-location and co-time are both learnable, with a BERT fine-tuned model outperforming several other baselines.
The MS-LaTTE dataset provides an opportunity to tackle many new modeling challenges in contextual task understanding and we hope that its release will spur future research in task intelligence more broadly.
 \\ \newline  }

\begin{document}

\maketitleabstract

\section{Introduction}\label{sec:intro}


Tasks are the primary unit of personal and professional productivity. People capture, organize, track and complete tasks as a way to measure and make progress towards their goals. Task management strategies range from scribbled sticky notes on refrigerators to complex group collaboration platforms such as \href{https://tasks.office.com}{Microsoft Planner}, and everything in between. Digital tools for task management support are prevalent in a range of applications including electronic mail~\cite{bellotti2003taking}, to-do applications~\cite{Bellotti2004WhatAT} and digital assistants~\cite{GrausEtAl2016}, and for a range of user scenarios such as contextual reminders~\cite{kamar2011jogger}, task duration estimation~\cite{white2019task}, and complex task decomposition~\cite{zhang2021learning}.


Many of these applications record and utilize a number of signals such as the text of a task, the time it was created, it's importance, due-date, and who it is assigned to, in order to build intelligent solutions for the user. However, they all notably assume tasks are homogeneous from the perspective of the \emph{completion} context. This assumption is flawed because tasks are, in fact, deeply context dependent; completing or making progress on tasks depends on extraneous factors, such as time in schedule, proximity to home or businesses, and resource availability. For example, a user is likely to want a reminder to buy eggs when they are close to a grocery store, as opposed to when they are at an airport.
 
Two types of context that are especially salient are location and time. These signals are readily accessible to systems and have been studied in previous work on contextual understanding~\cite{GrausEtAl2016,Bellotti2004WhatAT,Benetka2019UnderstandingCF}, recommendation~\cite{zhuang2011recommendation,YaoEtAl2015,zeng2016online} and reminders~\cite{kamar2011jogger}. However, there continues to remain a disconnect between logged contextual signals and actions taken in the real world, since users often record completion of tasks at a later time and a different location~\cite{zhang2022task}. Moreover, most prior work leverages proprietary data and does not make these available to other researchers for further study. The lack of a sizeable publicly available task dataset, and especially one that is tagged with location and time meta-data, has limited research on task intelligence in general, and particularly on the important area of contextual task modeling.

Thus, in this paper we are releasing a novel resource called the \emph{Microsoft Locations and Times of Task Execution} (MS-LaTTE) dataset. This dataset of 10,101 tasks sourced from real-world data is the largest publicly available repository of to-do tasks of any kind, and an order of magnitude larger than previously collected datasets~\cite{Landes2018ASA}. Additionally, it is the only dataset that also contains contextual location and time labels for where and when the tasks are likely to be completed. The dataset is available \href{https://github.com/microsoft/MS-LaTTE}{here}.


In addition to collecting and releasing the data, we also explore MS-LaTTE in this paper, to assess its utility for contextual task modeling. Specifically, we analyze the annotations to see if they capture interesting properties or regularities that might be useful for downstream modeling. Additionally, we motivate and establish two new benchmark evaluation tasks derived from the MS-LaTTE dataset -- Co-location and Co-time prediction -- and evaluate a number of popular language modeling approaches against these benchmarks. We find that learning is possible on both benchmarks, and a fine-tuned BERT approach significantly outperforms other baselines. However, both Co-location and Co-time prediction are difficult problems, and allow for more complex modeling efforts in future work to outperform the systems that we evaluate.

In summary, we make the following contributions in this paper:
\begin{enumerate}[topsep=2pt,itemsep=-1ex,partopsep=1ex,parsep=1ex]
    \item Highlight the opportunities and challenges in contextual task modeling.
    \item Compile and publicly release a novel dataset with over 10k tasks (an order of magnitude larger than previously available datasets) labeled with location and time meta-data (\S\ref{sec:collection}).
    \item Perform a detailed analysis of the annotations, demonstrating that they capture intuitively reasonable regularities (\S\ref{sec:analysis}).
    \item Conduct a modeling experiment to demonstrate the viability of the data for machine learning applications, setting up the novel benchmark tasks of Co-location and Co-time prediction in the process (\S\ref{sec:modeling}).
    \item Present future directions, including further applications and additional contexts (\S\ref{sec:conclusion}).
\end{enumerate}


\section{Related Work}\label{sec:related}
Previous investigations in a few distinct areas are relevant to the research described in this paper, and include work on task management, context-aware computing, and tasks data.

\subsection{Task Management} Task management systems that assist people in successfully managing, prioritizing, structuring, organizing and completing tasks have been the subject of a large body of research. There are many applications and assistants on the market today, including Amazon Alexa, Google Assistant, todoist, Trello, Microsoft To-Do, to name only a few.
It has long been recognized that task management does not take place in isolation but rather in the rich context of daily life. Hence research on task management and assistants is situated within the broader research area of context-aware computing. Location and time are two important contextual factors for both personal task assistance~\cite{Benetka2019UnderstandingCF,10.1145/2638728.2641718} and work-related task management~\cite{940025,Rhodes97}. Even before the existence of smaller computing devices such as cellphones that allow location-aware computing, actions as simple as setting an alarm or time-based reminder make it abundantly clear that there is a strong need for temporal context in successful task management. 

\newcite{Bellotti2004WhatAT} point out that co-location of tasks minimizes the need for multiple excursions for a user and that certain kinds of tasks show effects of periodicity, in terms of certain days, weeks and time of year. \newcite{GrausEtAl2016} study reminder data from the Cortana personal assistant. They establish that different task types cluster around different user-selected notification times. Communication task reminders, for example, tend to have a notification time that falls into typical work hours, chore notifications are more commonly set to early morning hours and tasks that involve moving to a specific location tend to have reminders set for lunch time at work. Interestingly in their data set, knowing the time at which a reminder was created is a stronger predictor of notification time than the task title itself. \newcite{LudfordFinnertyEtAl} highlight the fact that not only location but also movement patterns detected by a cellphone are important. By combining location with time information, context-aware task management systems can better assist users by suggesting the right tasks at the right time~\cite{kessell2006castaway,Rhodes97}. However, the role of context in task management is still only partially understood.

\subsection{Context-Aware Computing} Beyond task management, in the wider arena of context-aware computing and recommendation systems, time and location signals are also fundamental.
Context-aware recommendation of points of interest (POI) is typically studied based on check-in data from social network applications such as Foursquare and uses location as well as time~\cite{YaoEtAl2015,Zhao_Zhu_Liu_Xu_Li_Zhuang_Sheng_Zhou_2019}. POI recommendation also has a strong social component, making it amenable to the use of collaborative filtering signals in addition to spatio-temporal context~\cite{YuanEtAl2013}. Other applications include generating online recommendations for advertising and news, based on temporal patterns~\cite{zeng2016online}, and recommending entities based on historic user behavior and spatio-temporal sensor context~\cite{zhuang2011recommendation}.

Other broad research areas where time and location (as well as social context and travel trajectory) provide important information are mobile search~\cite{TeevanEtAl2011,10.1145/2207676.2208644}, opportunistic routing (driving assistance)~\cite{horvitz2012some}, and information retrieval in general~\cite{bennett2011inferring,radinsky2013behavioral}.

\subsection{Tasks Data} Despite the extensive research in task management and assistance, there is very little publicly available data. Typically, the data sources used in the literature are proprietary and subject to strict privacy requirements. To the best of our knowledge, \newcite{Landes2018ASA} are the only researchers to have publicly released a dataset of to-do tasks. It consists of approximately 600 tasks provided by users of \href{https://trello.com}{Trello} who volunteered their data, as well as from public Trello boards. The tasks are annotated for
\begin{enumerate*}[(a)]
    \item an intelligent agent that would be appropriate to assist with the task (the set of 15 agents serve as a task taxonomy), and
    \item the argument(s) of the action expressed in the task
\end{enumerate*}. 

In comparison, MS-LaTTE is the first public large-scale dataset of real-world to-do tasks, consisting of more than 10k instances. It is also the only dataset to contain annotated meta-data about the locations and times at which tasks are likely to be completed.
\section{Dataset Collection}\label{sec:collection}

In this section, we describe the annotation setup for collecting the MS-LaTTE dataset. Our goal is to source real-world tasks and capture the locations and times at which they are usually completed.

\subsection{Sourcing Real-world To-do Data}\label{subsec:rwdata}

We source tasks for our dataset from a sample of the logs of the now-defunct \href{https://en.wikipedia.org/wiki/Wunderlist}{Wunderlist} application. These logs are only obtained after a thorough legal- and trust-approved enterprise-grade pipeline processes them to anonymize and scrub all personally identifiable information. In addition, the pipeline performs k-anonymization so that tasks that were created by fewer than five users or fewer than 100 times in total are automatically discarded. 
The result is an \emph{aggregate} view of the logs, devoid of any identifiers, private information or infrequent tasks that can be correlated back to a user. What remains is a collection of task titles (such as ``buy milk'', ``mow the lawn'' etc.) along with list titles to which they are commonly assigned by users (such as ``groceries'', ``home chores'', etc.).

This aggregate view provides a rich collection of real-world to-dos from which we sample items for annotation. Unfortunately, grocery -- and to a lesser extent packing -- related tasks are over-represented in the aggregate data, accounting for over 70\% of distinct items, by our estimates. Therefore, rather than sample uniformly at random, we use heuristics to under-sample grocery and packing tasks. This is to avoid a resulting annotated dataset where a large part is trivially assigned a single location (or time) label. Our heuristic uses a manually curated set of the most popular list titles from the data related to groceries and packing (e.g. ``grocery'', ``safeway'', ``packing list''), and samples task titles from these lists at a lower rate than from other lists. Specifically we use an 10-10-80 percentage sampling probability for grocery, packing, and other lists respectively.

In this manner, we sample a total of 12,000 distinct task-list pairs
that are subsequently annotated with location and time information.

\subsection{Annotation Setup}\label{subsec:annsetup}

We used an internal crowd-sourcing platform to delegate work to non-expert workers in India contracted to perform annotations
These workers are payed a fixed hourly rate 
rather than an amount per HIT completed, which in our experience, disincentivizes cheating and has led to better annotation quality. We conducted the annotation over the sampled collection of task-list pairs in two distinct stages: first for location, then for time. This ordering was motivated by a couple of initial pilot annotation rounds that demonstrated that annotators agreed on location to a much higher degree than on time; this makes sense, since tasks are much more likely to be completed at the same location by different users than at the same time (due to individual schedules and preferences).
While the annotation stages are broadly similar, there are a few important differences that we highlight in what follows.

\begin{figure*}
\begin{center}
\includegraphics[scale=0.75]{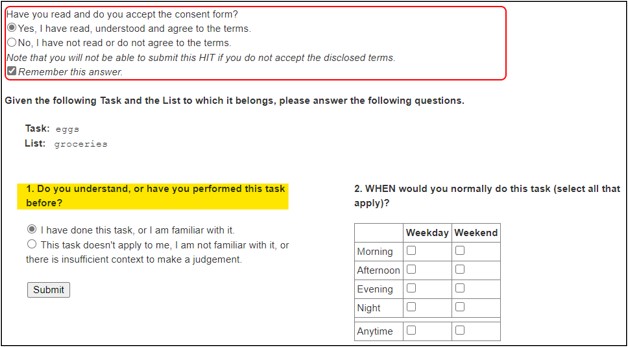}
\caption{The interface used for collecting location and time for task completion. Question 2 differs between stages of annotation; in this example an annotator is asked to provide time labels}
\label{fig:setup}
\end{center}
\end{figure*}

Annotators are presented with an interface like the one in Figure~\ref{fig:setup}, where the unit for a HIT is a single task-list pair. After providing their consent to have their annotations collected for research and machine learning,
annotators are first asked whether they are familiar with or have performed the task in the task-list pair before. If they answer in the negative, the HIT is considered complete and they are allowed to move on to the next one. This is consistent across both stages of annotation. If, however, they answer in the affirmative, they are then asked to provide labels for location or time. Specifically, they are asked {\bf WHERE} or {\bf WHEN} they would normally do this task, and they are permitted to select more than one label.

For location, annotators are provided with four broad categories from which to select:
\begin{enumerate*}[(a)]
    \item Home
    \item Work
    \item A public location
    \item Somewhere else (along with a free-form text box) 
\end{enumerate*}
Note that our guidelines broadly specify that ``Work'' can cover a number of locations, depending on a task creator's likely primary vocation, including ``school'' or ``college'' for students. Further, if (c) is selected, annotators are asked to specify at least one of several public location labels (e.g., ``grocery store'', ``dentist'', ``library'' etc.). Initially, a list of 36 such public locations were manually curated from a taxonomy of map location categories. Then, over a few initial pilot rounds of annotation, inputs from option (d) are used to refine this list into a final set of 69 public location labels.
For annotator convenience, public locations are manually organized into seven broad categories such as ``retail'', ``recreation'', ``finance'' etc. -- these are only shown in the annotation interface, and do not form part of the final dataset.

It should be noted that annotators were instructed to respond by providing labels for \emph{physical} locations of task completion. That is, even if a task may be completed online, the labels reflect the physical locations at which it is normally completed. This was done to reduce confusion and ambiguity, as well as to make the data more directly grounded in the real world and usable by future applications that can leverage geo-location information.
Tasks that are completed online are nevertheless a very interesting area of research that we hope to explore in future work.

Annotations were collected between May and August 2020, during the COVID-19 pandemic, when boundaries between \emph{Home}, \emph{Work} and other public locations were sometimes blurred\footnote{For example, ``mow the lawn'', ``exec review'' and ``haircut'' are tasks that can all be legitimately assigned to \emph{Home} in the present circumstances, whereas before the pandemic this might not be the case.}.
To account for this, annotators were instructed to answer questions using pre-COVID times as a contextual frame of reference.

In the first stage, each of the 12,000 HITs was labeled by three annotators. Those HITs that were marked as unfamiliar or unknown to two or more annotators were discarded from the dataset. Furthermore, in instances where all three annotators did not have a single location label in common, the first author of this paper acted as a fourth annotator;
365 HITs were thus supplementally annotated. The few remaining instances where there was four-way disagreement on location labels were also subsequently discarded. In the end, a total of 1,899 task-list pairs from the original set of 12,000 were removed, leaving 10,101 tasks.

The remaining tasks were annotated for time in the second stage of annotation. Annotators were asked to select one or more time labels from a set of 10 time buckets (as shown in Figure~\ref{fig:setup}), each specified by two distinct dimensions: the time of the day, and the day of the week. Times of day included:
\begin{enumerate*}[(a)]
    \item Morning
    \item Afternoon
    \item Evening
    \item Night, and
    \item Anytime;
\end{enumerate*}
while days of the week could be either:
\begin{enumerate*}[(i)]
    \item Weekday, or
    \item Weekend.
\end{enumerate*}
Annotators were instructed
to interpret the different times of day according to their own frame of reference (for example \emph{Morning} might mean 5am-8am to an early riser, but 8am-11am for someone else), thereby allowing for label alignment across annotators and tasks at a conceptual level rather than according to strictly defined time buckets. Additionally, they were told to use \emph{Anytime} when they were no more likely to complete a task at any one of the other times of day.

As with location labeling, annotators were asked to consider their pre-pandemic schedules and propensities for doing tasks, when judging the data.

Since {\bf WHEN} someone completes a task is altogether more subjective than {\bf WHERE}, it is expected that there will be a great deal more variation in labels for the second stage of annotation. In fact, it would be more appropriate to call time annotation a \emph{survey} rather than a labeling task. To account for and capture some of this variability we asked five annotators to label each of the 10,101 task-list pairs in the dataset, and none of the responses were discarded. Of course, even five annotations is insufficient to fully capture the breadth of preferences for when tasks are completed; however, the dataset we have collected allows for easy extension of time information with more responses in future work.

In summary, the final MS-LaTTE dataset -- collected over two stages of annotation -- consists of 10,101 real-world task-list pairs, each of which includes a set of labels for location from three annotators\footnote{222 instances in the dataset contain judgements from four rather than three annotators, for the reasons noted above.},
and for time from five annotators.
To the best of our knowledge, it is the largest (by an order of magnitude) dataset of to-do like tasks sourced from real users, and the only one to contain explicit contextual information in the form of locations and times at which tasks are completed.
\section{Analysis of the Dataset}\label{sec:analysis}

In this section, we describe an analysis of the MS-LaTTE dataset. We begin by measuring agreement between annotators. We then investigate the distributions of labels across location and time independently, and what annotator consensus looks like for each. Finally, we examine the relationship between location and time labels by performing a cross-correlation exploration of the data.


\subsection{Annotator Agreement}\label{subsec:agreement}

We use Krippendorff's Alpha~\cite{krippendorff2011computing} to measure the degree of annotator agreement. Since annotators can provide multiple labels to instances in the dataset, we apply the MASI~\cite{passonneau2006measuring} distance metric, a measure of agreement over set-valued objects. This setup yields agreement values of 0.50 and 0.09 for Location and Time respectively, which in turn are considered moderate and poor degrees of inter-rater agreement~\cite{mchugh2012interrater}.

As noted in \S\ref{subsec:annsetup}, we hypothesized that labels on Time are subjective and therefore the low annotator agreement is expected.
The ability for annotators to provide multiple labels for each instance also negatively impacts the inter-rater reliability metric. If singleton labels -- that is, those that were provided only by one annotator for a given instance -- are removed, Krippendorff's Alpha values become 0.87 and 0.26, which are excellent and fair for location and time respectively. In other words, while human judges may have individual preferences for where or when they expect to complete tasks, they are more likely to agree on a core set of locations or times at which these tasks should be completed. This is important, not only to contextualize the agreement numbers, but because it means that we can leverage the existence of this core agreement set for predictive modeling.



\subsection{Label Distribution}

\begin{figure*}[ht!]
\centering
\captionsetup[subfigure]{width=0.95\linewidth}
\begin{subfigure}{.48\textwidth}
  \centering
  \includegraphics[width=.95\linewidth]{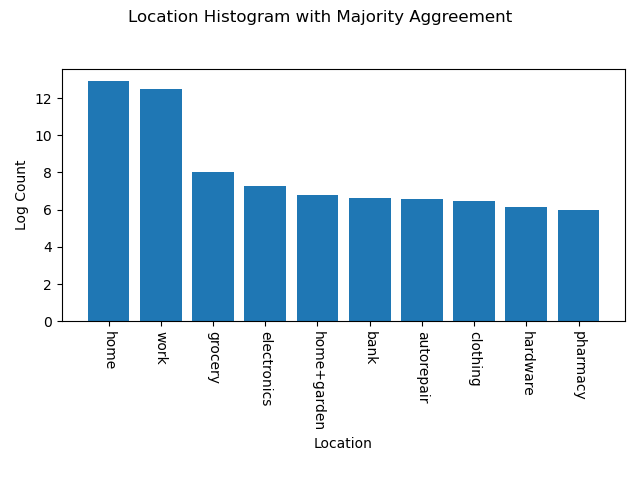}
  \caption{A histogram (counts in log base 2) of the 10 most popular location labels, where the majority of annotators agreed on the label.}
  \label{fig:histo:loc}
\end{subfigure}
\begin{subfigure}{.48\textwidth}
  \centering
  \includegraphics[width=.95\linewidth]{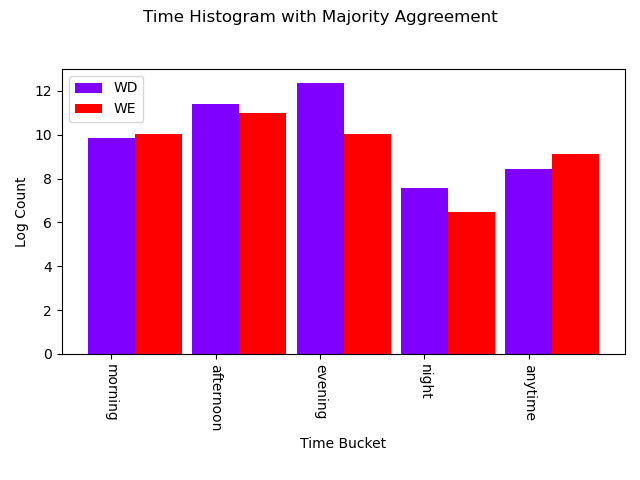}
  \caption{A histogram (counts in log base 2) of time bucket labels, where the majority of annotators agreed on the time bucket label.}
  \label{fig:histo:tim}
\end{subfigure}
\caption{Histograms of majority agreement labels over the Location and Time annotations in the dataset. The y-axes are presented in log (base 2) of the raw counts to accentuate differences between labels.}
\label{fig:histo}
\end{figure*}

We now turn to the analysis of the dataset itself and we begin by providing a snapshot view of the label distributions over location and time portions of the data, in the form of histograms. They are shown in Figures~\ref{fig:histo:loc} and~\ref{fig:histo:tim} respectively. Note that log counts (base 2) are used instead of raw counts to show differences between labels more clearly. We plot the histogram over labels on which there was majority agreement between annotators; this is to mitigate any impact that singleton labels (as noted in \S\ref{subsec:agreement}) may have on observable trends.

Figure~\ref{fig:histo:loc} demonstrates that \emph{home} and \emph{work} are by far the most popular locations for tasks in the dataset, and are an order of magnitude more frequent than any of the other labels. \emph{Grocery} stores are the third most frequent label, despite the heuristics that we used to under-sample grocery related tasks (\S\ref{subsec:rwdata}). The rest of the labels, corresponding to purchase or errand related locations (and including the remaining 60 not shown on the figure), follow a long-tailed distribution. These general trends align with our observations about the larger aggregate Wunderlist logs, which forms the basis for the MS-LaTTE dataset; namely that grocery, home, and work-related tasks form the overwhelming majority of tasks that users like to track in to-do related apps.

Meanwhile, Figure~\ref{fig:histo:tim} also displays some interesting trends over the distribution of time labels. Weekday evenings appear to be the most active time for completing tasks, perhaps due to the likelihood of users possibly being at \emph{home}, \emph{work} or running errands (like at a \emph{grocery} store), where many tasks are frequently completed. The popularity of \emph{home} or \emph{work} related tasks is also potentially a contributing factor to why mornings and afternoons are additional common times for completing tasks. A more in-depth analysis on cross-correlations between location and time labels is presented in \S\ref{subsec:crosscorr}.

The difference between weekdays and weekends also evidences some interesting properties. While people are generally less active completing tasks on weekends (sometimes starkly, such as in the evenings), they are actually slightly more active on weekend mornings -- perhaps due to home chores or other errands that are set aside specifically for non-working days. Additionally, tasks that are categorized as being done \emph{anytime} (often very short, simple tasks that require little planning, as we show in \S\ref{subsec:consensus}) are more likely to be completed on weekends than on weekdays. This is perhaps due to weekends providing more leisure time for unplanned tasks.

\subsection{Label Consensus}\label{subsec:consensus}

\begin{table*}[ht!]
\resizebox{\textwidth}{!}{
\begin{tabular}{ccccccc}
\hline 
{\bf home} & {\bf work} & {\bf office supply} & {\bf pharmacy} & {\bf electronics store} & {\bf clothing store} & {\bf hardware store}\tabularnewline
\hline 
rearrange closet & meeting tasks & buy envelopes & dr refill & bring in headphone & office attire & get a tape measure\tabularnewline
fix tv remote & sociology paper & buy sharpies & zzzquil liquid & dad speakers & astronaut costume & pressure washer part\tabularnewline
put on license sticker & finish udemy course & get packing tape & pick up relpax & more usb cables & scouts uniform & make house key copy\tabularnewline
\hline 
\end{tabular}
}
\caption{Examples of tasks that annotators commonly agreed should be assigned to specific location categories.}
\label{tabl:loctasks}
\end{table*}

\begin{table*}[ht!]
\resizebox{\textwidth}{!}{
\begin{tabular}{ccccccc}
\hline
{\bf WD morning} & {\bf WD afternoon} & {\bf WD evening} & {\bf WD anytime} & {\bf WE morning} & {\bf WE afternoon} & {\bf WE evening}\tabularnewline
\hline 
antibiotic morning  & meeting tasks  & bring book in & call sumo & antibiotic morning & platinum caulk & return items \tabularnewline
live class & sort out direct debits & pickup drycleaner & phone to mummy & finish gym & paint toilet & target vitamins for week\tabularnewline
get breakfast and coffee & work on newsletter & upload instagram & email to pronob & water poinsettias & lunch with parents & change tank filter\tabularnewline
\hline 
\end{tabular}
}
\caption{Examples of tasks that annotators commonly agreed should be completed during a specific time bucket.}
\label{tabl:timtasks}
\end{table*}

Our computation of annotator agreement by discarding singleton labels (\S\ref{subsec:agreement}), seemed to indicate that annotators tend to agree on core sets of labels for specific tasks, especially in the case of location. Thus, we now provide examples of high-agreement labels for both location and time, to show that the information provided by annotators captures reasonable location and temporal expectations for task completion. Tables~\ref{tabl:loctasks} and~\ref{tabl:timtasks} give some examples of tasks that were assigned the same label by a majority of annotators; not all labels are included in either table due to space constraints.

Table~\ref{tabl:loctasks} demonstrates that location labels are often incontrovertible and that annotators are able to correctly agree on the most likely location for tasks, despite the more than 70 possible labels to choose from. Meanwhile, Table~\ref{tabl:timtasks} also shows that when the majority of annotators agree on a time bucket for a task, they select labels that are very reasonable. In other words, while individual people may choose to complete some of these example tasks at different times, due to personal preference or schedule, when several judges agree on a time bucket for a task, it appears to be a label that is easily interpretable -- for example, work tasks on weekday afternoons, or errands and hobbies in the evenings. Additionally, as previously noted, the \emph{anytime} label is often associated with very short tasks that do not require prior planning.

\begin{table*}[ht!]
\resizebox{\textwidth}{!}{
\begin{tabular}{cccc|ccc}
\hline 
\multicolumn{4}{c|}{{\bf Location}} & \multicolumn{3}{c}{{\bf Time}}\tabularnewline
\hline 
home & home+garden store & grocery store & doctor's office & WD morning & WE afternoon & WD anytime\tabularnewline
\hline 
laundromat & hardware store & pharmacy & hospital & WD afternoon & WE evening & WE anytime\tabularnewline
beauty salon & sporting goods store & home+garden store & pharmacy & WD evening & WD evening & WD afternoon\tabularnewline
work & grocery store & home & home & WE morning & WE morning & WD morning\tabularnewline
\hline 
\end{tabular}
}
\caption{Examples of location and time labels that were commonly co-assigned by annotators, as measured by point-wise mutual information.}
\label{tabl:colablpmi}
\end{table*}

While tables~\ref{tabl:loctasks} and~\ref{tabl:timtasks} provide agreement on single labels, a reasonable follow-up question is to ask whether multi-label annotations also capture useful signal. We attempt to answer this question by calculating the point-wise mutual information (PMI) between location and time labels independently, when these labels are applied to the same task by annotators. Intuitively, the PMI captures the degree to which labels are co-assigned, thereby providing a way to assess the interpretability of multi-label annotations.

Table~\ref{tabl:colablpmi} provides examples of such co-assignments. Each column gives the top three labels (by PMI) that co-occur with the label in the column header. Note that items within a column are unrelated, and their only shared connection is co-occurrence with the column header. Only a subset of location and time labels are provided, for brevity. As can be seen from the table, co-assignment by annotators often makes intuitive sense. For example, laundry can be done both at home or at a laundromat, home \& garden stores sell similar goods to hardware stores, grocery stores and pharmacies are co-located in the same building, and short unplanned tasks can be done anytime on weekdays or weekends.

\subsection{Location and Time Cross-Correlation}\label{subsec:crosscorr}

\begin{figure*}
\captionsetup[subfigure]{width=0.9\linewidth}
\centering
\begin{subfigure}{.5\textwidth}
  \centering
  \includegraphics[width=.9\linewidth]{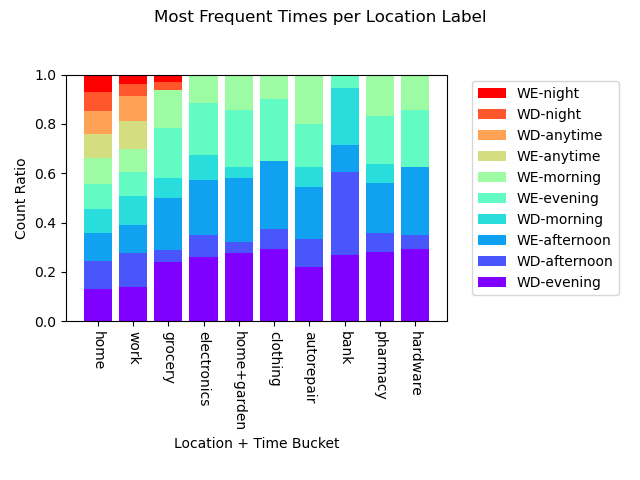}
  \caption{The distribution of time bucket labels over the 10 most popular location categories.}
  \label{fig:loc+tim:sub1}
\end{subfigure}%
\begin{subfigure}{.5\textwidth}
  \centering
  \includegraphics[width=.9\linewidth]{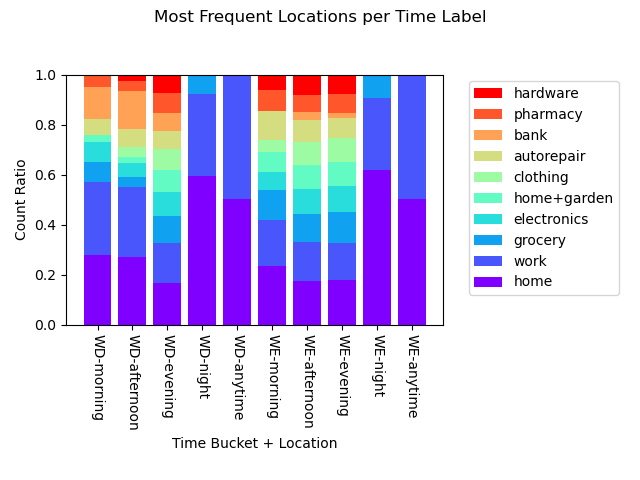}
  \caption{The distribution of the 10 most popular location labels over time buckets.}
  \label{fig:loc+tim:sub2}
\end{subfigure}
\caption{Stack plots showing the distribution over Location and Time label pairs}
\label{fig:loc+tim}
\end{figure*}

Up to this point, our analyses have looked at the location and time parts of the dataset separately, but since each of the 10,101 tasks in the dataset are annotated with both types of labels we can also conduct an analysis that looks at them jointly. Specifically, we can investigate whether location labels assigned to tasks make intuitive sense with respect to time, and vice-versa. Stack plots attempting to answer the question of cross-correlation between label sets are presented in Figures~\ref{fig:loc+tim:sub1} and~\ref{fig:loc+tim:sub2}.

For each pair of location category and time bucket, we count the number of tasks that were assigned both labels (each by majority agreement). Then we marginalize over time and location to get the stacked bars in Figures~\ref{fig:loc+tim:sub1} and~\ref{fig:loc+tim:sub2}, respectively. Note that we only use the 10 most popular location labels in this analysis.

These plots reveal some interesting and intuitively sensible findings. For example, in Figure~\ref{fig:loc+tim:sub1}, \emph{home} and \emph{work} are the only two locations where tasks are completed in every time bucket\footnote{It may be surprising that a number of \emph{work} tasks are completed on weekends, but recall this label is vocation dependent, and includes locations such as college (\S\ref{subsec:annsetup}).}; this makes sense since people spend the most time at these two locations. Another interesting observation is the relative proportion of tasks done on weekends at different public locations. For example, far more \emph{grocery}, \emph{home \& garden} or \emph{hardware} tasks are completed on weekends than \emph{bank} tasks,
since it is the expectation that the former are open for business while the latter are not (or may only be open with limited hours).

Figure~\ref{fig:loc+tim:sub2} also contains some noteworthy details. One example is that tasks assigned \emph{anytime} labels are only done at \emph{home} or at \emph{work}; this makes a lot of sense, considering that they are typically short, unplanned ones (\S\ref{subsec:consensus}), and that tasks completed at any other public location typically requires some form of forethought or planning. Another example, is the relative proportion of \emph{work} related tasks completed on different days of the week (especially in the mornings and afternoons), which conforms to the expectation that most people work on weekdays, rather than on weekends.

\begin{table}
\centering
\small
\begin{tabular}{cc}
\hline 
\textbf{Location} & \textbf{Time}\tabularnewline
\hline 
restaurant & WD+WE night\tabularnewline
dmv & WD afternoon\tabularnewline
movie theater & WE evening\tabularnewline
library & WD+WE night\tabularnewline
car wash & WE morning\tabularnewline
gym & WE morning\tabularnewline
\hline 
\end{tabular}
\caption{Examples of some location and time labels that were highly correlated, as measured by point-wise mutual information.}
\label{tabl:crosslablpmi}
\end{table}

Figures~\ref{fig:loc+tim:sub1} and~\ref{fig:loc+tim:sub2} present a broad picture of cross-correlation between location and time labels, but it may be useful to also explore a more focused view of cross-correlation. Table~\ref{tabl:crosslablpmi} presents some examples of the most highly correlated pairs of labels, as measured by PMI. This value is computed from probabilities of independent and joint label occurrences, which are obtained from counts over the location and time labels that received majority agreement. Because PMI is sensitive to very infrequent events, we discarded location labels that have fewer than five tasks associated with them. As can be seen from the table, these highly correlated label pairs make intuitive sense: for example, the fact that \emph{restaurant} and \emph{library} are associated with the \emph{night} time bucket on both weekdays and weekends (i.e., date nights or study sessions), that \emph{gym} is associated with \emph{WE morning} (i.e., early workout on weekends), or that \emph{dmv} is associated with \emph{WD afternoon} (i.e., less busy times when people are often at work\footnote{See: \href{https://www.ncdot.gov/dmv/offices-services/locate-dmv-office/Pages/helpful-hints.aspx}{helpful hints at www.ncdot.gov}})

\vspace{1mm}

In summary, this section has presented an analysis of the MS-LaTTE dataset. Our main takeaway from this analysis is the fact that while the data does contain some expected variance -- in the form of individual latitude for times at which tasks are completed -- interesting and often intuitively reasonable properties of task completion are captured by annotations with majority agreement. This motivates the use of MS-LaTTE for the learning effort we tackle in \S\ref{sec:modeling}, as well as for future work that we hope will leverage the dataset for modeling task intelligence.
\section{Modeling Co-Location and Co-Time}\label{sec:modeling}


Given the annotations in the dataset, there are many interesting predictive problems that can be tackled such as predicting the location or time bucket most likely to support some task activity, or predicting when (resp. where) a task should happen given a user's set location (resp. time bucket) and description, or even scheduling a user's day using their time commitments and likely locations. However, in this paper we seek only to describe and validate the MS-LaTTE dataset and therefore tackle the foundational modeling problems of predicting whether two tasks are likely to be completed together (by location or time); other modeling efforts are left to the community and to future work.

This leads us to the two benchmark tasks of co-location prediction and co-time prediction. Note that while these tasks may be simpler than some of the more ambitious modeling challenges described above, a model that successfully tackles the former may indicate approaches that might succeed on the latter. Moreover, co-location prediction and co-time prediction are meaningful modeling efforts in and of themselves, leading to potential user-facing scenarios such as alerting users working on a given task, to other tasks they can complete, based on their current location or time (or both).

Formally given two tasks $\mathcal{T}_1 = (t_1, l_1)$ and $\mathcal{T}_2 = (t_2, l_2)$, where $t$ and $l$ are task and list descriptions respectively, the problems of co-location and co-time are to find binary predictive functions $f(\mathcal{T}_1, \mathcal{T}_2) \rightarrow {0, 1}$. We generate benchmark datasets for evaluating these two predictive tasks from the annotated MS-LaTTE dataset. We sample 25,000 task pairs from the cross product of the 10,101 unique tasks in MS-LaTTE, using 20,000, 1,000, and 4,000 respectively for training, validation, and test splits. Pairs are assigned a positive label if they contain at least one common label that was assigned by a majority of annotators; otherwise they are assigned a negative label.

While pairs are unique, tasks themselves may repeat. To ensure fair evaluation we stratify the dataset so that tasks that appear in any split do not appear in any other split. It may be noted that the resulting benchmark datasets are imbalanced containing a roughly 71/29 and 38/62 positive/negative split for location and time respectively. The train, validation, and test splits for co-location and co-time benchmarks are released with the dataset. In this paper, we report accuracy and Macro F1 (due to the imbalance in the datasets) as evaluation metrics for the two evaluation tasks.

We evaluate several popular language modeling approaches in this paper. They include:
\begin{enumerate*}[(a)]
    \item {\bf Random} -- a baseline that randomly assigns a positive or negative label to an instance, using the ratios in the training data as bases for sampling a label.
    \item {\bf Lexical} -- a model that featurizes both sets of tasks and list strings and uses uni-, bi- and tri-gram features in a logistic regression classifier trained and tuned on the train and validation splits respectively.
    \item {\bf GloVe} -- a model that uses the popular GloVe vectors~\cite{pennington2014glove} as features for lexical items in the task and list strings. An average of the GloVe vectors is passed through a logistic classifier that is trained and tuned on the train and validation splits respectively.
    \item {\bf BERT} -- a model that is similar to the GloVe model above, but uses pre-trained BERT embeddings~\cite{vaswani2017attention} instead. Because these representations capture full strings as opposed to individual tokens, the embeddings for $\mathcal{T}_1$ and $\mathcal{T}_2$ are concatenated rather than averaged.
\end{enumerate*}

Finally, we compare these models, against a more sophisticated fine-tuned BERT model. Like the pre-trained BERT model, this one also featurizes both task inputs $\mathcal{T}_1$ and $\mathcal{T}_2$. But it uses a composition variant that has been successfully applied in past work to problems that involve dual string comparison~\cite{mou2015natural}, such as textual entailment:

\begin{equation}\label{eq:concat}
    c = [\mathcal{T}_1; \mathcal{T}_2; \mathcal{T}_1 - \mathcal{T}_2; \mathcal{T}_1 \circ \mathcal{T}_2]
\end{equation}

\noindent where the semi-colon represents concatenation, and the $\circ$ signifies the element-wise dot product. In our model ({\bf BERT TE-FT}) the representation $c$ is then passed through a non-linear layer, before a final linear layer with a sigmoidal output produces a binary prediction value. The model uses an intermediate dimension of 256 as an output to a ReLU non-linearity; it also applies a dropout factor of 0.5 to the model during training to avoid overfitting. We use binary cross-entropy as the loss function, and we train the model using an Adam optimizer with a fixed weight decay~\cite{loshchilov2017decoupled}, and initial hyperparameters of $lr = 1e^{-5}$, $eps=1e^{-8}$.

\begin{table}[ht!]
\resizebox{0.5\textwidth}{!}{
\begin{tabular}{c|cc|cc}
\hline 
 & \multicolumn{2}{c|}{Co-location} & \multicolumn{2}{c}{Co-time}\tabularnewline
\hline 
 & Accuracy & Macro F1 & Accuracy & Macro F1\tabularnewline
\hline 
Random & 59.00 & 0.501 & 51.85 & 0.492\tabularnewline
Lexical & 77.38 & 0.696 & 57.47 & 0.518\tabularnewline
GloVe & 74.48 & 0.613 & \textbf{60.40} & 0.498\tabularnewline
BERT & 75.40 & 0.625 & 59.28 & 0.511\tabularnewline
BERT TE-FT & \textbf{81.98} & \textbf{0.773} & 59.10 & \textbf{0.569}\tabularnewline
\hline 
\end{tabular}
}
\caption{Results of several models on the binary prediction tasks of co-location and co-time detection.}
\label{tabl:results}
\end{table}

The results of our evaluation are given in Table~\ref{tabl:results}. They demonstrate that while all models are capable of outperforming the random baseline, they do so with varying degrees of success. The best model is the BERT TE-FT model, which outperforms the other approaches by a
significant margin on both benchmark tasks\footnote{It does not achieve the highest accuracy on co-time prediction due to the imbalanced nature of the dataset.}. These improvements are statistically significant at a p-value=0.01 based on a paired student's t-test. The lexical model proves surprisingly capable, outperforming both GloVe and BERT models; this is possibly because the training data size is fairly large and contains a decent coverage of lexical terms, negating the need for the fuzzy matching afforded by fixed embeddings. An interesting avenue for future work is to compare models on varying amounts of training data. Notably, both tasks are challenging (with co-time prediction being significantly more so), and there is likely room for improvement over the simple approaches described in this paper.

\begin{table}[ht!]
\resizebox{0.5\textwidth}{!}{
\begin{tabular}{c|cc|cc}
\hline 
 & \multicolumn{2}{c|}{Co-location} & \multicolumn{2}{c}{Co-time}\tabularnewline
\hline 
 & Accuracy & Macro F1 & Accuracy & Macro F1\tabularnewline
\hline 
BERT TE-FT & \textbf{81.98} & \textbf{0.773} & 59.10 & \textbf{0.569}\tabularnewline
(-) Fine-tuning & 71.58 & 0.426 & \textbf{61.45} & 0.381\tabularnewline
(-) TE Composition & 80.90 & 0.757 & 56.70 & 0.541\tabularnewline
\hline 
\end{tabular}
}
\caption{An ablation study demonstrating the effects of removing components of the full BERT TE-FT model.}
\label{tabl:ablation}
\end{table}

While BERT TE-FT is clearly the best model on both benchmark tasks, we can also attempt to evaluate how much contribution its distinctive features add by conducting an ablation study. Specifically, we ablate the model by:
\begin{enumerate*}[(a)]
    \item freezing the parameters of the BERT model (effectively negating the effects of fine-tuning); or
    \item using a simple concatenation of $\mathcal{T}_1$ and $\mathcal{T}_2$ instead of the more complex vector composition variant in Equation~\ref{eq:concat}.
\end{enumerate*}
The results of our ablation study are given in Table~\ref{tabl:ablation}. They show that while both components influence the full model positively, fine-tuning is clearly a far more important positive factor. This seems to indicate that the language used in to-do tasks is different from general purpose text on which BERT is trained, and thus benefits from fine-tuning to the domain.

\vspace{1mm}

In conclusion, we presented the two new benchmark tasks of co-location prediction and co-time prediction, derived from the MS-LaTTE dataset. We compared a number of popular language modeling approaches on these benchmarks and showed that they can indeed be modeled successfully. Notably, a model containing fine-tuned BERT seemed to perform best. However, both benchmarks are challenging and present future work with interesting possibilities to outperform the simple models in this paper with more sophisticated approaches.
\section{Conclusion and Future Work}\label{sec:conclusion}

We have publicly released a new dataset of to-do tasks called MS-LaTTE. This dataset contains location and time labels from multiple annotators for every one of its 10,101 tasks, and is the first to contain such contextual information surrounding task completion. It is also the largest publicly available dataset of real-world to-do tasks of any kind, by an order of magnitude. In this paper, we have described the setup used to collect and annotate the dataset, conducted a detailed analysis of its labels and properties, and performed experimental evaluations on two novel benchmark tasks -- co-location prediction and co-time prediction -- derived from the dataset. We found that the data captures several intuitive regularities, and that these regularities can be modeled by popular language modeling techniques -- including, most successfully, by a BERT fine-tuned approach. We anticipate that the release of MS-LaTTE will spur the research community to work on contextual task modeling, and more generally on task intelligence.

Despite the utility, we hope the community will derive from this dataset, it does have some limitations. We rely on third-party judges' interpretations of tasks that they did not create themselves, and for which they have very little information besides the raw textual representation. Moreover, annotators are all from a single locale (India) and thus may miss broader cultural or country-specific subtleties that are outside their field of experience. These issues may have potentially led to labels that are erroneous or noisy due to misconstrued intent.

To resolve these and other issues, there are several interesting and challenging research directions that we hope to pursue in future work. They include:
\begin{enumerate*}[(a)]
    \item alternative mechanisms to gather context information for tasks directly from individuals, such as in-situ data collection and experience sampling methods;
    \item leveraging contexts beyond time and place, such as people, activity, busyness and resources required (digital as well as physical);
    \item using MS-LaTTE to model novel contextual prediction and recommendation scenarios, such as predicting when or where a task is most likely to be completed, ranking tasks according to their likelihood of being completed given location, time or both, or recommending other tasks based on what a user is currently doing, where and/or when;
    \item personalization of models to individuals or cohorts, based on factors such as occupation, preferences and habits;
    \item and integration of models in real-world task management systems and digital assistants, with user studies and online experimentation to determine downstream impact on users' productivity.
\end{enumerate*}

\section{Bibliographical References}\label{reference}

\bibliographystyle{lrec2022-bib}
\bibliography{bibliography}


\end{document}